%% file: main.tex
\PassOptionsToPackage{table}{xcolor}
\PassOptionsToPackage{numbers,square,sort&compress}{natbib}
\documentclass[11pt]{article}

\usepackage[preprint]{acl}

\usepackage{times}
\usepackage{latexsym}
\usepackage[utf8]{inputenc}

\IfFileExists{arabic.ldf}{%
  \usepackage[LAE,T1]{fontenc}%
  \usepackage[arabic,main=english]{babel}%
}{%
  \usepackage[T1]{fontenc}%
}

\usepackage{microtype}
\usepackage{graphicx}
\usepackage{booktabs}
\usepackage{float}
\usepackage{multirow}
\usepackage{amsmath}
\usepackage{tikz}
\usetikzlibrary{positioning,arrows.meta}
\usepackage{pgfplots}
\pgfplotsset{compat=1.17}
\usepackage{siunitx}
\input{macros}

\title{\asr{}: A Production Multidialectal {A}rabic Speech Recognition System}

\author{%
  Mahmoud Khaled, Mohammed Abd-Elnaser, Mahmoud Mohsen, \\
  \textbf{Hazem Ahmed, Abdelrahman Hafez} \\
  EchoNova AI, Dubai, United Arab Emirates \\
  \texttt{\{mahmoud, mabdelnasser, mmohsen, hazem, abdelrahman\}@echonova.ai} \\[6pt]
  {\small\href{https://github.com/echonova-ai-solutions/arabic-asr-benchmark}{\linkicon{github}Code}%
  \qquad
  \href{https://huggingface.co/datasets/echonova-ai/arabic-dialect-asr-benchmark}{\linkicon{hf}Dataset}%
  \qquad
  \href{https://echonova.ai}{\linkicon{globe}echonova.ai}}
}

\begin{document}
\maketitle

\input{sections/00_abstract}
\input{sections/01_introduction}
\input{sections/02_related}

\input{sections/03_system}
\input{sections/04_data_pipeline}
\input{sections/05_methodology}
\input{sections/07_asr_eval}

\input{sections/08_error_analysis}
\input{sections/09_conclusion}
\input{sections/10_limitations}
\input{sections/11_ethics}

\bibliography{refs}

\onecolumn
\appendix
\setcounter{section}{0}
\renewcommand{\thesection}{\Alph{section}}
\renewcommand{\thesubsection}{\thesection.\arabic{subsection}}
\input{sections/A_appendix}

\end{document}

%% file: sections/00_abstract.tex
\begin{abstract}
Callers to Arabic customer-service lines speak their own dialect, while speech
recognizers are built and evaluated mainly on Modern Standard Arabic (MSA). A
recognizer that mishears the dialect breaks the call before anything else can
help. We present \asr{}, the speech recognizer of EchoNova's production Arabic
voice agent. On a 120-hour test set of MSA and eight dialects, it has the
lowest word and character error rates of twelve systems, against eleven from
Microsoft, ElevenLabs, Google, OpenAI, Speechmatics, IFM, Cohere, NVIDIA and
Audar: 11.69\% word error rate against
13.51\% for the strongest, Microsoft's MAI-Transcribe-2 (a lead of 0.74 points
on clips unaffected by that system's fixed number and script conventions), and
15.52\% against 16.96\% averaged over varieties. Its largest margins are on
Omani (4.30 points), Egyptian (3.78) and Saudi (2.19) speech; it is level on
Syrian and loses on Lebanese (24.96\% against 23.05\%), mostly because it
writes Lebanese words in MSA forms and is less verbatim. On six public test
sets its average word error rate under the leaderboard's current scoring is
17.17\%, below every published figure we found, among them Falcon-ASR
(20.92\%), Jais-ASR (22.85\%) and Audar-ASR-V1-Turbo (23.17\%), though none
of these states that it used that scoring, so the
comparison is indicative. We describe the data pipeline and methodology
behind \asr{} and release a 20-hour subset of the test set with our scoring
code.
\end{abstract}

%% file: sections/01_introduction.tex
\section{Introduction}
\label{sec:intro}

Arabic is diglossic \citep{ferguson1959diglossia}: Modern Standard Arabic (MSA)
is the language of writing, news and official documents, while everyday speech
happens in regional dialects that differ from MSA and from each other in
vocabulary, pronunciation and morphology. A caller to an Arabic customer-service line
speaks their dialect, not MSA. Dialects have no standard orthography
\citep{habash2012coda}, and speakers mix in English words
\citep{hamed2020arzen}. Speech technology built and evaluated mainly on MSA
meets this speech at its weakest point.

Recognition is where that mismatch does its damage first: a voice agent
mishearing a caller's dialect is wrong before the rest of the system has a
chance to be right, and the caller cannot see the transcript to correct it.

We present \asr{},\footnote{Sada (\ar{صدى}) means `echo'. The name is unrelated
to \sadacorpus{} \citep{alharbi2024sada}, one of the public test sets we report
on.} the speech recognizer of \sys{}, a production Arabic voice agent for
customer service, and evaluate it against eleven commercial and open-weight
systems on a 120-hour private test set of MSA and dialects from the Gulf, Egypt and the
Levant. \asr{} has the lowest word and character error rates, with its largest
margins over the strongest external system, Microsoft's MAI-Transcribe-2, on
Omani, Egyptian and Saudi speech. It loses on Lebanese to MAI-Transcribe-2,
and we analyse why. On six public test sets, under the leaderboard's current
scoring, its average error rate is also below every published figure we found,
including those for Falcon-ASR and Jais-ASR; none of those figures
states that it used this scoring, so the comparison is indicative.

Our contributions are:
\begin{itemize}\setlength{\itemsep}{1pt}\setlength{\parskip}{0pt}
  \item \textbf{The data pipeline and methodology behind \asr{}}
    (Sections~\ref{sec:data} and~\ref{sec:method}).
  \item \textbf{A head-to-head comparison of twelve systems} on one private
    120-hour multi-dialect test set, all measured by us (Section~\ref{sec:asr-eval}).
  \item \textbf{Per-variety results}, including the one dialect where \asr{}
    loses and an analysis of why it loses there
    (Sections~\ref{sec:asr-results} and~\ref{sec:asr-errors}).
  \item \textbf{A comparison on six public test sets} with published results
    for 24 systems (Section~\ref{sec:asr-public}).
  \item \textbf{A release}: a 20-hour subset of the test set covering all nine
    varieties, with its reference transcripts, and our scoring code.
\end{itemize}

%% file: sections/02_related.tex
\section{Related Work}
\label{sec:related}

\paragraph{Arabic speech recognition and its benchmarks.} The MGB challenges
evaluated Arabic ASR first on largely MSA broadcast speech \citep{ali2016mgb2}
and then on dialectal web video \citep{ali2017mgb3,ali2019mgb5}, and QASR added about
\num{2000} hours of multi-dialect broadcast speech \citep{mubarak2021qasr}.
\citet{hussein2022arabic} benchmark end-to-end and modular recognizers on the
MGB-2, MGB-3 and MGB-5 test sets and compare them with human transcribers on a new one, and \citet{talafha2023nshot} benchmark Whisper
across Arabic varieties. Crowd-sourced and web-scale
corpora followed \citep{ardila2020commonvoice,alfetyani2023masc}, as did corpora
for single regions such as \sadacorpus{} \citep{alharbi2024sada}, multi-dialect
collections such as Casablanca \citep{talafha2024casablanca}, and corpora of
Arabic--English code-switching \citep{hamed2020arzen}, which multilingual
recognizers also target \citep{chowdhury2021towards}. The Open Universal
Arabic ASR Leaderboard now evaluates six of these test sets
\citep{wang2024openuniversal}, and NADI 2025 added a multidialectal ASR track
\citep{talafha2025nadi}. Among industrial systems, \citet{salhab2025advancing}
train on automatically labelled speech and report leaderboard results against
open-weight models and commercial services, and \citet{salhab2025nadi} extend the approach to the NADI
dialects; multilingual models such as Whisper \citep{radford2022whisper} serve
as common baselines. Our evaluation differs in what it compares and how: current
commercial APIs and open-weight models measured by us on one private test set,
with results for each variety.

\paragraph{Orthography and scoring.} Dialectal Arabic has no standard
orthography; CODA and its unified successor propose conventions for annotating
it \citep{habash2012coda,habash2018codastar}. Single-reference WER therefore
penalizes valid spelling variants, which \citet{ali2015mrwer} address by
scoring against multiple references.

%% file: sections/03_system.tex
\section{Deployment Context}
\label{sec:system}

\begin{figure*}[t]
\centering
\begin{tikzpicture}[
  font=\small,
  box/.style={draw, rounded corners=2pt, align=center, minimum height=0.95cm, inner sep=4pt},
  arr/.style={-{Stealth[length=2mm]}, thick},
  node distance=0.55cm and 0.5cm]
  \node[box] (caller) {Caller\\\scriptsize phone / web};
  \node[box, right=of caller] (vad) {Voice activity\\detection};
  \node[box, right=of vad] (asr) {\asr{}};
  \node[box, right=of asr] (llm) {LLM\\\scriptsize tools};
  \node[box, right=of llm] (tts) {TTS};
  \draw[arr] (caller) -- (vad);
  \draw[arr] (vad) -- (asr);
  \draw[arr] (asr) -- (llm);
  \draw[arr] (llm) -- (tts);
  \draw[arr] (tts.south) |- ++(0,-0.7) -| (caller.south);
\end{tikzpicture}
\caption{Where \asr{} sits in the voice agent at call time.}
\label{fig:system}
\end{figure*}
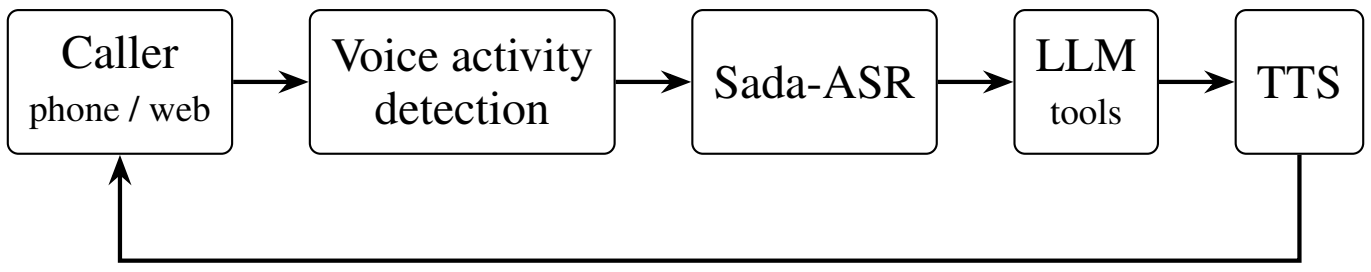

\asr{} is deployed inside \sys{}, a cascaded voice agent for customer
service: a speech recognizer, a hosted large language model (LLM) with tools,
and a text-to-speech (TTS) model, connected through a real-time media server
(Figure~\ref{fig:system}). A cascade was chosen over an end-to-end speech model
because each speech stage can then be specialized to Arabic dialects on its
own, the LLM can be replaced without retraining either speech model, and every
intermediate result is text that can be checked before it is spoken.

The TTS model produces 24\,kHz speech and was trained on \num{20000} hours of
speech. It provides voices in Emirati, Kuwaiti, Egyptian, Saudi and Modern
Standard Arabic, and in English; each agent is configured with a subset of
them. The agent and its TTS model are the setting \asr{} is built for; neither
is evaluated in this report.

%% file: sections/04_data_pipeline.tex
\section{Data Pipeline}
\label{sec:data}

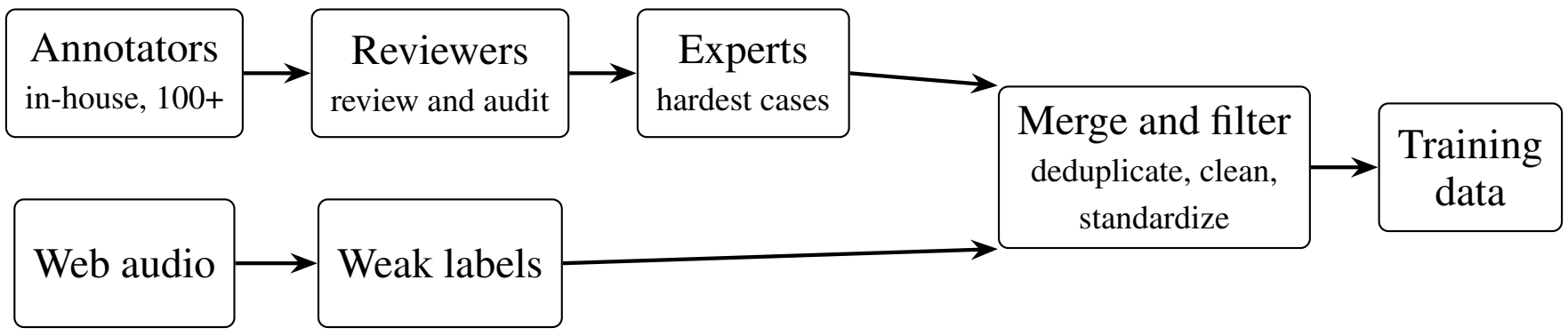
\begin{figure*}[t]
\centering
\begin{tikzpicture}[
  font=\small,
  box/.style={draw, rounded corners=2pt, align=center, minimum height=0.95cm, inner sep=4pt},
  arr/.style={-{Stealth[length=2mm]}, thick},
  node distance=0.45cm and 0.5cm]
  \node[box] (ann) {Annotators\\\scriptsize in-house, 100+};
  \node[box, right=of ann] (rev) {Reviewers\\\scriptsize review and audit};
  \node[box, right=of rev] (exp) {Experts\\\scriptsize hardest cases};
  \node[box, below=of ann] (web) {Web audio};
  \node[box, below=of rev] (weak) {Weak labels};
  \node[box, right=1.1cm of exp, yshift=-0.7cm] (merge) {Merge and filter\\\scriptsize deduplicate, clean,\\\scriptsize standardize};
  \node[box, right=of merge] (train) {Training\\data};
  \draw[arr] (ann) -- (rev);
  \draw[arr] (rev) -- (exp);
  \draw[arr] (web) -- (weak);
  \draw[arr] (exp.east) -- (merge.north west);
  \draw[arr] (weak.east) -- (merge.south west);
  \draw[arr] (merge) -- (train);
\end{tikzpicture}
\caption{The two data streams behind \asr{}: audio transcribed on our annotation
platform, and weakly labelled web audio, merged into one training set.}
\label{fig:data}
\end{figure*}

Verbatim transcripts of dialectal Arabic are scarce. Each dialect needs
annotators who speak it, and dialects have no standard spelling to transcribe
against. Dialectal audio is easier to find than its transcripts, but the labels
that come with web audio are weak. \asr{} is therefore trained on two streams
of data, merged into one training set (Figure~\ref{fig:data}): audio
transcribed on our own annotation platform, and weakly labelled audio collected
from the web. The corpus also contains English speech, about a third of all
training hours.

\subsection{In-House Annotation Platform}
\label{sec:data-platform}
We built an annotation platform with three layers. More than 100 in-house
annotators from across the Arab world transcribe the audio. Reviewers continuously review and audit the
annotators' submissions. Experts, who are professional writers and domain
specialists, handle the hardest cases.

The platform measures annotators and reviewers with control samples. The same
evaluation samples are placed in every annotator's queue, which gives each
annotator's performance a common baseline. Duplicate samples are inserted
without the annotators' knowledge, which measures how consistently different
annotators transcribe the same audio. All control samples are removed before
the dataset is compiled.

\subsection{Weakly Labelled Web Data}
\label{sec:data-web}
To add volume and real-world variety, we collected thousands of hours of weakly
labelled audio from the web.

\subsection{Merging and Filtering}
\label{sec:data-merge}
The two streams differ in label quality and in format, so they are merged in
three steps:
\begin{itemize}\setlength{\itemsep}{1pt}\setlength{\parskip}{0pt}
  \item \textbf{Deduplication.} Repeated audio is removed across the two
    streams.
  \item \textbf{Cleanup.} Corrupted records, mismatches between audio and
    label, and audio artifacts are filtered out.
  \item \textbf{Standardization.} Audio formats and metadata are made uniform
    across both sources, which yields one training set.
\end{itemize}

\begin{figure}[t]
\centering
\begin{tikzpicture}
\begin{axis}[
  xbar, bar width=6pt,
  scale only axis, width=0.55\columnwidth, height=3.4cm,
  xmin=0, xmax=56,
  /pgf/number format/assume math mode=true,
  axis x line=none, axis y line*=left, y axis line style={draw=none},
  symbolic y coords={MSA,Egyptian,Saudi,Syrian,Emirati,Kuwaiti,Omani,Lebanese,Qatari},
  ytick=data,
  y dir=reverse, enlarge y limits=0.05,
  yticklabel style={font=\small}, tick style={draw=none},
  nodes near coords={\pgfmathprintnumber[fixed,precision=1,zerofill]{\pgfplotspointmeta}\%},
  nodes near coords style={font=\scriptsize, color=inksecondary, anchor=west, xshift=1pt},
  clip=false,
]
\addplot[fill=seriesblue, draw=none] coordinates {
  (45.0,MSA) (19.5,Egyptian) (17.7,Saudi) (6.9,Syrian) (4.6,Emirati)
  (2.9,Kuwaiti) (1.5,Omani) (1.2,Lebanese) (0.7,Qatari)};
\end{axis}
\end{tikzpicture}
\caption{Relative share of \asr{}'s dialect-labelled Arabic training audio
across the nine varieties of our test set.}
\label{fig:data-mix}
\end{figure}
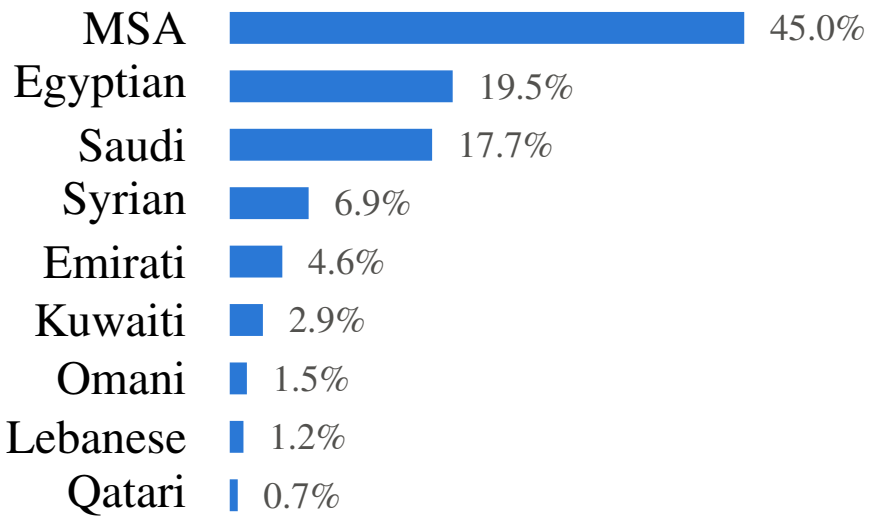

Figure~\ref{fig:data-mix} shows how the training audio divides across the
varieties of our test set. Within these nine varieties, MSA is just under half
of the training audio, Egyptian about a fifth and Saudi about a sixth; Omani,
Lebanese and Qatari are each under 2\%.

%% file: sections/05_methodology.tex
\section{Methodology}
\label{sec:method}

\subsection{Architecture}
\label{sec:method-arch}
\asr{} is an attention-based encoder--decoder recognizer that writes the
transcript token by token.

\subsection{Training}
\label{sec:method-train}
We train \asr{} on the corpus of Section~\ref{sec:data}. Each epoch
draws half of its clips from English, repeating English audio and subsampling
Arabic; within Arabic, clips are drawn in proportion to each variety's hours
(Figure~\ref{fig:data-mix}). We apply no data augmentation: no added noise,
speed perturbation, codec simulation, reverberation or SpecAugment. Audio is
only resampled to 16\,kHz mono.

The objective is next-token cross-entropy on the transcript and its
end-of-sequence token, with no label smoothing and no auxiliary CTC loss. Arabic
targets are undiacritized, keep the hamza forms of alif, write numbers as
words, and keep code-switched words in Latin script, lowercased; English
targets are cased and keep digits. Punctuation and casing in the targets come
from an automatic punctuation-restoration model.

We select checkpoints by word error rate on a held-out validation subset of
\num{1482} utterances (at most 100 per variety, English included), decoded
with plain greedy search. The subset is disjoint from the training clips and from every test
set we report.

\subsection{Decoding}
\label{sec:method-decode}
Every test set, private and public, is decoded with the same configuration:
plain greedy search, with no beam search, repetition penalty or n-gram
constraint, and no decode-time tuning. A guard stops repetition loops, allowing at most three repeats of any 4- to
8-token unit, and each
window's output is capped at 14 tokens per second of audio plus 16 (at least
20). Audio longer than 30\,s is decoded in 30\,s windows with 2\,s overlap and
the window texts are joined. Clips are decoded one at a time.

%% file: sections/07_asr_eval.tex
\section{Speech Recognition Evaluation}
\label{sec:asr-eval}

We compare \asr{} with eleven external recognizers---five commercial APIs and
six self-hosted open-weight models---on a private test set
(Sections~\ref{sec:asr-data}--\ref{sec:asr-results}). The test set was
unreleased when the systems were measured, the APIs of Microsoft, Google and
OpenAI do not train on submitted audio by default, and the open-weight models ran on our own
hardware. No clip of the private test set, and no part of the six public test
splits, was used to train or select \asr{}. Six public test sets connect our
numbers to published results for 24 systems (Section~\ref{sec:asr-public}).

\subsection{Private Multi-Dialect Test Set}
\label{sec:asr-data}

The test set contains 120 hours of speech in \num{53597} clips covering nine
varieties: Modern Standard Arabic (MSA) and eight regional dialects
(Table~\ref{tab:asr-corpus}). It is stratified by duration---each variety
contributes a fixed number of hours---so clip counts vary with speaking rate and
segmentation. Every clip is at most 30\,s long, so no system needed long-form
segmentation and all received identical audio. The clips are cut from
\num{14686} source recordings; clips from one recording share speaker and
channel, so our confidence intervals resample recordings.

MSA is 41.6\% of the audio and our aggregate is a corpus-level micro-average,
so the overall figure is dominated by MSA: a system that is strong on MSA and
weak on dialects still looks good in aggregate. We therefore treat the
per-variety results as primary.

\begin{table}[t]
\centering
\small
\setlength{\tabcolsep}{4pt}
\begin{tabular}{lrrrrr}
\toprule
 & \multicolumn{3}{c}{Test set} & \multicolumn{2}{c}{Released} \\
\cmidrule(lr){2-4}\cmidrule(lr){5-6}
Variety & Hours & Clips & Mean (s) & Hours & Clips \\
\midrule
MSA       & 49.8 & \num{20452} & 8.8 & 8.27 & \num{3400} \\
Egyptian  & 20.2 & \num{9049}  & 8.0 & 3.37 & \num{1508} \\
Emirati   & 15.0 & \num{7119}  & 7.6 & 2.50 & \num{1187} \\
Saudi     & 10.0 & \num{5492}  & 6.6 & 1.67 & 916 \\
Kuwaiti   &  5.0 & \num{2535}  & 7.1 & 0.84 & 423 \\
Qatari    &  5.0 & \num{2378}  & 7.6 & 0.83 & 396 \\
Omani     &  4.8 & \num{2226}  & 7.8 & 0.81 & 371 \\
Syrian    &  5.0 & \num{2166}  & 8.3 & 0.83 & 361 \\
Lebanese  &  5.0 & \num{2180}  & 8.3 & 0.83 & 363 \\
\midrule
Total     & 119.9 & \num{53597} & 8.1 & 19.95 & \num{8925} \\
\bottomrule
\end{tabular}
\caption{Composition of the private test set and of its released subset.
Mean is the mean clip length; totals are computed before rounding.}
\label{tab:asr-corpus}
\end{table}

\subsection{Systems, Configuration and Scoring}
\label{sec:asr-systems}

Table~\ref{tab:asr-main} lists the systems.
All systems were measured on 13 October 2026, except Jais-ASR and
Speechmatics, measured on 15--16 October. Every API was set to Arabic, and
to verbatim output with formatting options disabled wherever the API exposes
them; no system---including ours---used contextual biasing, custom
vocabulary or any other adaptation. Microsoft's service was given the locale
\texttt{ar}; its API reports the locale \texttt{ar} whatever is sent, and two
dialect locales gave identical transcripts on 48 clips from six varieties. We selected Microsoft's model explicitly, ran ElevenLabs Scribe v2 with
audio-event tagging and diarization off, and ran Speechmatics with its enhanced
model, which keeps disfluencies and writes numbers as digits.
OpenAI's and Google's models were reached through OpenRouter at
temperature~0; Google's verbatim mode, which OpenRouter does not expose, is its
default. The open-weight models ran at their published or default
settings.\footnote{Open-weight models by Hugging Face ID:
\texttt{IFM/Jais-ASR},
\texttt{audarai/Audar-ASR-V1-Turbo},
\texttt{nvidia/\allowbreak stt\_ar\_\allowbreak fastconformer\_\allowbreak hybrid\_\allowbreak large\_pc\_v1.0},
\texttt{openai/whisper-large-v3},
\texttt{nvidia/\allowbreak nemotron-3.5-asr-\allowbreak streaming-0.6b} and
\texttt{CohereLabs/\allowbreak cohere-transcribe-\allowbreak arabic-07-2026}.
Whisper was decoded greedily with Arabic forced, Nemotron in streaming mode at
its 1120\,ms setting, FastConformer with its default decoder, Jais-ASR with its model card's
decoding (greedy, repetition penalty 1.05) and Arabic set, Audar with Arabic
set, and Cohere Transcribe greedily with Arabic set.}

WER and CER are corpus micro-averages: total edit distance divided by total
reference length, in words or characters. On the private test set,
hypotheses and references are normalized identically: punctuation and diacritics are removed; the hamza and
madda forms of alif are folded into bare alif, hamza on waw and ya is folded
into the bare letter, and standalone hamza is deleted; the extended letters for
/p/ and /v/ are mapped to \ar{ب} and \ar{ف}; alif maqsura is not folded into ya;
Eastern Arabic digits are mapped to Western
digits; and \ar{ة} is folded into \ar{ه}, a common Arabic
normalization step \citep{obeid2020camel}. Numbers are not
converted between digits and words. For every system, any one- to three-word
unit repeated more than ten times in a row is cut to ten copies (no reference
repeats one more than seven times), and whitespace is canonicalized. The crop
changes four systems' WER (Gemini from 19.95\% to 16.13\%, Whisper from
28.94\% to 26.57\%, Jais-ASR from 16.53\% to 16.41\% and Cohere Transcribe
from 29.12\% to 29.11\%). An API output with more than five times the
reference's word count and more than 50 words was requested again, up to three
times, and the first normal output kept. Whisper, Audar, Nemotron and Jais-ASR
were decoded in batches, which is not bit-identical to decoding one clip at a
time; we did not measure the effect. For every
system and clip we store the raw output before normalization, so the reported
error rates can be recomputed without re-running a recognizer.

\begin{table*}[t]
\centering
\small
\begin{tabular}{llrrrr}
\toprule
System & Type & WER & Macro & CER & Empty \\
\midrule
\ours \asr{} (ours)                 & self-hosted & \best{11.69} & \best{15.52} & \best{4.76} & 0.00 \\
Microsoft MAI-Transcribe-2          & API         & 13.51 & 16.96 & 6.37  & 0.00 \\
ElevenLabs Scribe v2                & API         & 16.12 & 20.67 & 7.54  & 0.08 \\
Google Gemini 3.5 Transcribe        & API         & 16.13 & 20.84 & 8.23  & 0.02 \\
OpenAI gpt-transcribe               & API         & 17.03 & 21.69 & 8.59  & 0.01 \\
Speechmatics Enhanced               & API         & 18.08 & 24.37 & 8.73  & 0.00 \\
IFM Jais-ASR                        & self-hosted & 16.41 & 21.00 & 6.89  & 0.00 \\
Audar-ASR-V1-Turbo                  & self-hosted & 20.86 & 25.17 & 9.97  & 0.00 \\
NVIDIA FastConformer (Arabic)       & self-hosted & 25.97 & 32.93 & 12.31 & 0.01 \\
OpenAI Whisper large-v3 \citep{radford2022whisper} & self-hosted & 26.57 & 32.56 & 15.08 & 0.00 \\
NVIDIA Nemotron-3.5 streaming 0.6B  & self-hosted & 28.11 & 36.35 & 11.43 & 0.10 \\
Cohere Transcribe Arabic            & self-hosted & 29.11 & 37.94 & 19.41 & 0.05 \\
\bottomrule
\end{tabular}
\caption{Private test set. WER and CER in \%,
corpus micro-averaged; Macro is the unweighted mean of the nine per-variety WERs in
Table~\ref{tab:asr-variety}; Empty is the \% of clips with an empty transcript.}
\label{tab:asr-main}
\end{table*}

\subsection{Results}
\label{sec:asr-results}

\paragraph{Overall.} \asr{} has the lowest WER (11.69\%) and CER (4.76\%) of the
twelve systems (Table~\ref{tab:asr-main}). Against the strongest external system,
MAI-Transcribe-2 (13.51\%), this is a 13.5\% relative WER reduction: a
difference of 1.83 points (95\% CI 1.72--1.93, bootstrap over source
recordings). Averaged over the nine varieties without weighting by duration,
the WERs are 15.52\% and 16.96\%.
MAI-Transcribe-2's API writes numbers as digits and English words in Latin
script, and neither can be configured off; 9.23\% of its clips contain digits
where the reference has none, against none of \asr{}'s. On the clips with no
digit, number word or Latin script in the reference or in either system's
output (the controlled clips), the margin is 0.74 points (95\% CI 0.65--0.83), and
0.49 when hesitation fillers are also removed.

\begin{table*}[t]
\centering
\footnotesize
\setlength{\tabcolsep}{2.2pt}
\resizebox{\textwidth}{!}{%
\begin{tabular}{lrrrrrrrrrrrrrr}
\toprule
& & \multicolumn{5}{c}{API} & \multicolumn{6}{c}{Self-hosted} & & \\
\cmidrule(lr){3-7}\cmidrule(lr){8-13}
Variety & \asr{} & MAI-2 & Scribe v2 & Gemini & gpt-tr. & Speechm. & Jais & Audar & FastConf. & Whisper & Nemotron & Cohere & $\Delta$ MAI-2 (95\% CI) & $\Delta$ ctrl. \\
\midrule
MSA      & \best{5.61}  & 6.99  &  7.80 & 8.03  & 7.63  & 7.82 & 7.83 & 11.04 & 12.05 & 14.66 & 12.35 & 15.33 & $+1.38$ {\scriptsize[1.25, 1.51]} & $+0.23$ \\
Egyptian & \best{9.48}  & 13.26 & 15.19 & 14.32 & 17.78 & 17.56 & 16.55 & 25.45 & 28.65 & 28.11 & 31.57 & 25.33 & $+3.78$ {\scriptsize[3.59, 3.98]} & $+1.70$ \\
Saudi    & \best{14.74} & 16.93 & 21.36 & 20.68 & 21.18 & 20.94 & 21.06 & 23.95 & 31.72 & 31.11 & 34.40 & 31.32 & $+2.19$ {\scriptsize[1.81, 2.63]} & $+1.57$ \\
Emirati  & \best{18.57} & 20.22 & 24.40 & 24.85 & 26.30 & 26.78 & 23.86 & 26.86 & 37.22 & 37.24 & 37.95 & 45.31 & $+1.65$ {\scriptsize[1.37, 1.91]} & $+0.77$ \\
Kuwaiti  & \best{14.29} & 14.83 & 18.24 & 19.21 & 19.55 & 33.84 & 19.84 & 23.22 & 35.48 & 31.02 & 35.71 & 41.54 & $+0.53$ {\scriptsize[0.17, 0.87]} & $+0.58$ \\
Qatari   & \best{15.82} & 17.02 & 21.72 & 20.73 & 22.15 & 22.32 & 20.32 & 22.96 & 35.33 & 31.95 & 33.89 & 44.02 & $+1.20$ {\scriptsize[0.48, 1.95]} & $+0.89$ \\
Omani    & \best{16.07} & 20.37 & 23.05 & 22.35 & 23.24 & 22.81 & 22.65 & 26.45 & 34.55 & 33.93 & 36.23 & 40.09 & $+4.30$ {\scriptsize[3.82, 4.80]} & $+4.55$ \\
Syrian   & 20.16 & \best{19.98} & 24.90 & 25.70 & 26.93 & 32.89 & 26.54 & 31.63 & 38.55 & 39.60 & 49.74 & 41.85 & $-0.17^{\dagger}$ {\scriptsize[$-$0.72, 0.40]} & $-0.46$ \\
Lebanese & 24.96 & \best{23.05} & 29.37 & 31.73 & 30.43 & 34.37 & 30.35 & 34.93 & 42.84 & 45.41 & 55.32 & 56.67 & $-1.91$ {\scriptsize[$-$2.47, $-$1.35]} & $-3.27$ \\
\bottomrule
\end{tabular}}
\caption{WER (\%) by variety on the private test set. MAI-2: Microsoft
MAI-Transcribe-2; Scribe v2: ElevenLabs Scribe v2; Gemini: Google Gemini 3.5
Transcribe; gpt-tr.: OpenAI gpt-transcribe; Speechm.: Speechmatics Enhanced;
Jais: IFM Jais-ASR; FastConf.: NVIDIA FastConformer (Arabic); Nemotron: NVIDIA
Nemotron-3.5 streaming 0.6B; Cohere: Cohere Transcribe Arabic.
$\Delta$ MAI-2: \asr{}'s margin over MAI-Transcribe-2 in points (positive =
\asr{} lower), with a 95\% CI from \num{2000} bootstrap resamples of source
recordings; $^{\dagger}$the only CI including zero. $\Delta$ ctrl.: the same margin
on clips with no digit, number word or Latin script in the reference or in
either system's output. Differences are computed before rounding. CER:
Appendix~\ref{app:asr-variety}.}
\label{tab:asr-variety}
\end{table*}

\paragraph{By variety.} \asr{} has the lowest WER on seven of the nine varieties
(Table~\ref{tab:asr-variety}). On Syrian it is level with MAI-Transcribe-2
(20.16\% against 19.98\%; the CI of the difference includes zero), and on
Lebanese it is behind by 1.91 points, a loss analysed in
Section~\ref{sec:asr-errors}. Its margins over MAI-Transcribe-2
are largest on Omani (4.30 points, 21.1\% relative), Egyptian (3.78) and
Saudi (2.19), followed by Emirati (1.65), MSA (1.38) and Qatari (1.20); on
Kuwaiti the margin is a narrow 0.53. On the controlled clips, every margin keeps its sign: the largest are Omani
(4.55), Egyptian (1.70) and Saudi (1.57), Syrian stays level, and the Lebanese
loss widens to 3.27 points ($\Delta$ ctrl., Table~\ref{tab:asr-variety}). By
CER, \asr{} is the lowest of the twelve systems on all nine varieties, Lebanese
included; on Syrian the CI of its CER difference from MAI-Transcribe-2
includes zero ($+0.03$, 95\% CI $-0.41$ to $0.45$). We do
not claim that \asr{} degrades less than other systems from MSA to dialect: the
unweighted mean of its eight dialect WERs is 2.99 times its MSA WER, against
2.61 for MAI-Transcribe-2.

\paragraph{Empty and runaway output.} An empty transcript is scored as a full
deletion, so it is already inside WER; we report it separately because it is a
distinct failure in a live call. \asr{} returned an empty transcript for 2 of
the \num{53597} clips; among the other systems the rate reaches 0.10\%. None of
its transcripts is a runaway output---more than five times the reference's word
count and more than 50 words---against 131 of Whisper's.

\subsection{Public Test Sets}
\label{sec:asr-public}

\begin{table*}[t]
\centering
\footnotesize
\setlength{\tabcolsep}{4pt}
\begin{tabular}{lrrrrrrr}
\toprule
System & SADA & CV 18.0 & MASC-c & MASC-n & Casab. & MGB-2 & Avg \\
\midrule
\ours \asr{} (ours) & 25.76 & 3.95 & 5.35 & 16.70 & 41.72 & 9.56 & \best{17.17} \\
Falcon-ASR \citep{tii2026falconasr} & -- & -- & -- & -- & -- & -- & 20.92 \\
Audar-ASR-V1-Turbo, leaderboard \citep{elm2026leaderboardspace} & 28.92 & 8.09 & 16.73 & 27.19 & 47.02 & 11.08 & 23.17 \\
Munsit \citep{salhab2025advancing} & 27.71 & 10.42 & 21.74 & 28.08 & 60.04 & 12.10 & 26.68 \\
\midrule
\multicolumn{8}{l}{\textit{As reported by IFM \citep{jaisasr2026}}} \\
Jais-ASR & 30.85 & 8.56 & 16.12 & 25.44 & 46.56 & 9.59 & 22.85 \\
Speechmatics Melia 1 & 34.74 & 5.19 & 16.23 & 25.89 & 46.41 & 12.61 & 23.51 \\
ElevenLabs Scribe v2 & 33.96 & 8.71 & 16.91 & 27.15 & 50.20 & 10.30 & 24.54 \\
Cohere Transcribe Arabic & 37.47 & 5.82 & 19.60 & 27.07 & 49.71 & 15.54 & 25.87 \\
omniASR-LLM-7B & 41.61 & 8.75 & 19.69 & 29.29 & 56.46 & 14.13 & 28.32 \\
Qwen3-Omni-30B-A3B & 44.82 & 11.46 & 21.47 & 30.85 & 62.55 & 13.09 & 30.71 \\
Qwen3-ASR-1.7B & 45.53 & 16.90 & 24.37 & 34.29 & 64.47 & 16.57 & 33.36$^{*}$ \\
Voxtral-Small-24B & 50.82 & 15.25 & 23.96 & 34.43 & 66.30 & 16.03 & 34.47 \\
\midrule
\multicolumn{8}{l}{\textit{Commercial, as reported by \citet{salhab2025advancing}}} \\
ElevenLabs Scribe & 49.44 & 28.27 & 31.93 & 41.23 & 63.77 & 25.68 & 40.05 \\
OpenAI GPT-4o-transcribe & 66.47 & 28.19 & 31.53 & 43.29 & 70.72 & 29.62 & 44.97 \\
Microsoft Azure STT$^{\ddagger}$ & 58.50 & 33.77 & 40.66 & 45.64 & 64.84 & 30.91 & 45.72 \\
\midrule
\multicolumn{8}{l}{\textit{Open-weight, as reported by \citet{salhab2025advancing}}} \\
NVIDIA Conformer-CTC large & 47.26 & 10.60 & 24.12 & 35.64 & 71.13 & 19.69 & 34.74 \\
Whisper large-v3 & 55.96 & 17.83 & 24.66 & 34.63 & 71.81 & 16.26 & 36.86 \\
SeamlessM4T large-v2 \citep{seamless2023} & 62.52 & 21.70 & 25.04 & 33.24 & 66.25 & 20.23 & 38.16 \\
Whisper large-v3-turbo & 60.36 & 25.73 & 25.51 & 37.16 & 73.79 & 17.75 & 40.05 \\
Whisper large-v2 & 57.46 & 21.77 & 27.25 & 38.55 & 71.01 & 25.17 & 40.20 \\
Whisper large & 63.24 & 26.04 & 28.89 & 40.79 & 72.18 & 24.28 & 42.57 \\
Whisper medium & 67.71 & 28.07 & 29.99 & 42.91 & 75.44 & 29.32 & 45.57 \\
Parakeet CTC 1.1B (concat) & 70.70 & 26.34 & 30.49 & 45.95 & 80.80 & 24.94 & 46.54 \\
Parakeet CTC 1.1B (universal) & 73.58 & 40.01 & 36.16 & 50.03 & 81.30 & 30.68 & 51.96 \\
Whisper small & 78.02 & 24.18 & 35.93 & 56.36 & 87.64 & 48.64 & 55.13 \\
\bottomrule
\end{tabular}
\caption{WER (\%) on the six public test sets; Avg is their unweighted mean. Rows other than ours are published figures: Audar-ASR-V1-Turbo is the leaderboard's best entry (10 October 2026), and Falcon-ASR reports only an average over the same six test lists. IFM measured Jais-ASR with the leaderboard's evaluation script, without stating its version, measured Melia 1 and Scribe v2 itself, and lists the other five from the leaderboard. ElevenLabs Scribe, GPT-4o-transcribe and Azure STT in the Salhab rows are earlier models than those we measured. $^{\ddagger}$Mean of Azure's Saudi and Jordanian models. $^{*}$As published by IFM; its six per-set values average 33.69. Our row uses the leaderboard's current scoring, revised on 27 July 2026; most published rows predate that revision, so the comparison is indicative (CER: Appendix~\ref{app:asr-public-current}).}
\label{tab:asr-public}
\end{table*}

Six public test sets from the Open Universal Arabic ASR Leaderboard
\citep{wang2024openuniversal} connect our results to published ones:
\sadacorpus{} \citep{alharbi2024sada}, Common Voice 18.0
\citep{ardila2020commonvoice}, the clean and noisy test sets of MASC
\citep{alfetyani2023masc}, MGB-2 \citep{ali2016mgb2} and Casablanca
\citep{talafha2024casablanca}. Table~\ref{tab:asr-public} compares \asr{} with
the published figures for these sets: fourteen systems reported by
\citet{salhab2025advancing}, among them their own model and three commercial
services; eight reported by IFM \citep{jaisasr2026}, among them its own
Jais-ASR and the commercial Speechmatics Melia 1 and ElevenLabs Scribe v2;
Falcon-ASR \citep{tii2026falconasr}; and Audar-ASR-V1-Turbo, the
leaderboard's best entry on 10 October 2026 \citep{elm2026leaderboardspace}. We score the six sets with
the leaderboard's own evaluation code; \asr{} was measured on 13--14 October
2026. The leaderboard revised its scoring on 27 July 2026, fixing punctuation
removal, attaching a standalone \ar{و} to the next word and collapsing extra
spaces, and its earlier
rows were not re-scored. Under the current scoring, \asr{}'s WER averaged over
the six sets is 17.17\% over \num{46757} clips, against 20.92\% reported for
Falcon-ASR, 22.85\% for Jais-ASR, 23.17\% for Audar-ASR-V1-Turbo on the
leaderboard, 23.51\% for Speechmatics Melia 1 and 24.54\% for ElevenLabs
Scribe v2 as measured by IFM, and 34.74\% for the strongest open-weight model
and 40.05\% for the strongest of the three commercial services in
\citet{salhab2025advancing}. None of these figures states
that it used the current scoring, and how they treated punctuation is not
stated, so the comparison is indicative.

%% file: sections/08_error_analysis.tex
\section{Error Analysis: Where \asr{} Loses}
\label{sec:asr-errors}

Lebanese is a clear loss (Table~\ref{tab:asr-variety}), and a broad one:
\asr{} makes more errors than MAI-Transcribe-2 on 55 of the 71 Lebanese source
recordings, and the gap holds across clip lengths and speaking rates. The two
systems make almost the same share of substitutions; \asr{} deletes more words
and MAI-Transcribe-2 inserts more (Table~\ref{tab:leb-errors}). A word-by-word
comparison with the reference shows two habits of \asr{}.

\begin{table}[H]
\centering
\small
\setlength{\tabcolsep}{4pt}
\begin{tabular}{lrrrr}
\toprule
 & \multicolumn{2}{c}{Lebanese} & \multicolumn{2}{c}{Syrian} \\
\cmidrule(lr){2-3}\cmidrule(lr){4-5}
 & \asr{} & MAI-2 & \asr{} & MAI-2 \\
\midrule
Substitutions & 16.71 & 16.56 & 15.06 & 15.36 \\
Deletions     &  7.06 &  4.45 &  3.61 &  2.29 \\
Insertions    &  1.19 &  2.04 &  1.49 &  2.34 \\
\midrule
WER           & 24.96 & \best{23.05} & 20.16 & 19.98 \\
CER           & \best{10.86} & 11.76 & 7.79 & 7.81 \\
\bottomrule
\end{tabular}
\caption{Levantine errors (\%). Substitutions, deletions and insertions are
shares of the reference words and sum to WER before rounding.}
\label{tab:leb-errors}
\end{table}

First, \asr{} writes Lebanese words in MSA forms. The references spell words
shared with MSA as they are pronounced, as in \ar{كتير} `a lot' and \ar{اكتر}
`more'; \asr{} writes them with their MSA consonants, \ar{كثير} and \ar{اكثر}.
Net of MAI-Transcribe-2's errors of the same kind, this costs 1.18 points, 0.88
of them from the \ar{كتير} family alone (\ar{كتير}, \ar{اكتر}, \ar{بكتير}). \asr{} also replaces Lebanese
function words with their MSA counterparts, as in
\ar{هيدا}\,$\rightarrow$\,\ar{هذا} `this' and
\ar{نحنا}\,$\rightarrow$\,\ar{نحن} `we'. It also drops the \ar{ي} of the
verb prefix, as in \ar{بيصير}\,$\rightarrow$\,\ar{بصير}, which is not an MSA
form but changes the person of the verb, from `he becomes' to `I become'. The
MSA forms are its
own (Table~\ref{tab:leb-confusions}): MAI-Transcribe-2 almost never writes
\ar{كتير} as \ar{كثير}. Others, such as \ar{متل}\,$\rightarrow$\,\ar{مثل}
`like' and \ar{انتي}\,$\rightarrow$\,\ar{انت} `you', are shared conventions
that cost both systems equally and do not move the gap. As a diagnostic,
counting \ar{ت} and \ar{ث}, \ar{د} and \ar{ذ}, \ar{ز} and \ar{ظ}, and \ar{ا}
and \ar{ق} as equal for both systems lowers the gap from 1.91 to 0.51 points
(95\% CI $-$0.11 to 1.12), which is no longer significant.

\begin{table}[t]
\centering
\small
\begin{tabular}{lrr}
\toprule
Reference\,$\rightarrow$\,output & \asr{} & MAI-2 \\
\midrule
\multicolumn{3}{l}{\textit{Mostly \asr{}}} \\
\ar{كتير}\,$\rightarrow$\,\ar{كثير} `a lot' & 265 & 0 \\
\ar{هيدا}\,$\rightarrow$\,\ar{هذا} `this'  & 126 & 28 \\
\ar{نحنا}\,$\rightarrow$\,\ar{نحن} `we'    & 121 & 24 \\
\ar{اكتر}\,$\rightarrow$\,\ar{اكثر} `more' & 96 & 1 \\
\midrule
\multicolumn{3}{l}{\textit{Shared}} \\
\ar{متل}\,$\rightarrow$\,\ar{مثل} `like'   & 94 & 93 \\
\ar{انتي}\,$\rightarrow$\,\ar{انت} `you'   & 70 & 69 \\
\ar{راح}\,$\rightarrow$\,\ar{رح} `will'    & 35 & 41 \\
\midrule
\multicolumn{3}{l}{\textit{Mostly MAI-Transcribe-2}} \\
\ar{سو}\,$\rightarrow$\,So `so'            & 0 & 44 \\
\ar{اممم}\,$\rightarrow$\,\ar{امم} (filler) & 6 & 66 \\
\bottomrule
\end{tabular}
\caption{Frequent Lebanese substitutions and how often each system makes them.}
\label{tab:leb-confusions}
\end{table}

Second, \asr{} is less verbatim. It omits false starts, repeated words and
short words that the references keep, such as \ar{اي} `yes' and \ar{لا} `no',
which costs 1.35 points net. MAI-Transcribe-2 gives back 1.46 points through
more insertions, among them hesitation fillers where the reference has none,
and through other substitutions, such as English words written in Latin
script. Fillers do not explain the loss: both systems pay for them, since
\asr{} drops them and MAI-Transcribe-2 often writes them with a different
number of letters, and removing them from both sides widens the gap to 2.50
points. Substitutions of word form cost a whole word in WER but little in CER,
which helps explain why \asr{} still has the lower Lebanese CER.

The same habits appear on Syrian, more weakly (Table~\ref{tab:leb-errors}).
MSA spellings of shared words cost 0.74 points there, and MSA function words
0.06, since words such as \ar{هيدا} and \ar{نحنا} are characteristically
Lebanese. Fewer one-letter errors and fewer insertions than MAI-Transcribe-2
offset them, and the two systems are level on Syrian. We plan to address the
Lebanese loss by training on more verbatim Levantine transcripts with a
consistent dialect spelling.

%% file: sections/09_conclusion.tex
\section{Conclusion}
\label{sec:conclusion}

We presented \asr{}, the Arabic recognizer of a production voice agent, with
the data pipeline and methodology behind it, and evaluated it against eleven
external systems on a 120-hour test set of MSA and dialects from the
Gulf, Egypt and the Levant. \asr{} has the lowest word and character error
rates, with its largest margins on Omani, Egyptian and Saudi speech, and loses
on Lebanese to MAI-Transcribe-2, mostly because it writes Lebanese words in
MSA forms and is less verbatim. On six public test sets, under the leaderboard's current scoring, its average
word error rate is below every published figure we found, including those for
Falcon-ASR and Jais-ASR, though none of those states that it used this
scoring, so the comparison is indicative. The recognizer runs inside a production voice
agent, which we describe as context rather than evaluate.

We plan to test output on non-speech input, measure turn latency, run a
listening test of the TTS model and evaluate the full agent.

%% file: sections/10_limitations.tex
\section*{Limitations}
\label{sec:limitations}

The private test set covers MSA and eight dialects, and MSA is 41.6\% of its
audio; Maghrebi, Iraqi, Sudanese and Yemeni Arabic are not represented.

%% file: sections/11_ethics.tex
\section*{Ethics Statement}
\label{sec:ethics}

\paragraph{Third-party services.} Evaluating hosted systems meant sending test
audio to their providers' APIs, directly or through OpenRouter.

\paragraph{Release.} A 19.95-hour subset of the private test set, \num{8925}
clips from all nine varieties (Table~\ref{tab:asr-corpus}), is released with
its reference transcripts at
\url{https://huggingface.co/datasets/echonova-ai/arabic-dialect-asr-benchmark}.
Its transcripts, segmentation and metadata are licensed under CC BY 4.0; the
audio is for research and evaluation only, with no commercial use or
redistribution, and access requires accepting these terms. Our normalization
and scoring code is released at
\url{https://github.com/echonova-ai-solutions/arabic-asr-benchmark}.

\paragraph{Conflict of interest.} The authors are employed by the company that
develops \asr{}, and we ran every system on the private test set ourselves.
We mitigate this by reporting where \asr{} loses, dating every external
measurement,
storing raw outputs so the reported error rates can be recomputed, and
releasing the scoring code.

%% file: sections/A_appendix.tex
\section{Speech Recognition: Supplementary Results}
\label{app:asr}

\subsection{CER by Variety}
\label{app:asr-variety}

\begin{table}[H]
\centering
\footnotesize
\setlength{\tabcolsep}{3pt}
\resizebox{\textwidth}{!}{%
\begin{tabular}{lrrrrrrrrrrrr}
\toprule
& & \multicolumn{5}{c}{API} & \multicolumn{6}{c}{Self-hosted} \\
\cmidrule(lr){3-7}\cmidrule(lr){8-13}
Variety & \asr{} & MAI-2 & Scribe v2 & Gemini & gpt-tr. & Speechm. & Jais & Audar & FastConf. & Whisper & Nemotron & Cohere \\
\midrule
MSA      & \best{2.08}  & 3.58  & 3.72  & 4.17  & 3.63  & 3.57  & 3.21  & 6.22  & 6.33  & 9.68  & 4.72  & 11.36 \\
Egyptian & \best{3.73}  & 6.56  & 7.33  & 7.31  & 8.79  & 8.17  & 6.67  & 11.62 & 11.92 & 14.49 & 12.49 & 15.52 \\
Saudi    & \best{5.61}  & 7.53  & 9.72  & 9.79  & 9.82  & 9.79  & 8.57  & 10.17 & 14.24 & 16.30 & 13.52 & 19.25 \\
Emirati  & \best{8.54}  & 9.44  & 11.60 & 13.56 & 14.83 & 13.97 & 10.83 & 13.24 & 18.83 & 22.02 & 16.73 & 31.22 \\
Kuwaiti  & \best{4.79}  & 5.43  & 6.69  & 8.03  & 7.95  & 17.56 & 7.20  & 9.11  & 15.91 & 15.46 & 12.72 & 27.13 \\
Qatari   & \best{7.56}  & 8.56  & 10.85 & 11.48 & 12.78 & 11.76 & 9.65  & 11.53 & 18.74 & 18.38 & 15.14 & 31.13 \\
Omani    & \best{7.77}  & 10.02 & 11.55 & 11.96 & 13.11 & 11.19 & 11.15 & 13.57 & 18.35 & 20.37 & 17.22 & 28.08 \\
Syrian   & \best{7.79}  & 7.81  & 10.43 & 11.92 & 12.49 & 15.51 & 10.64 & 13.64 & 17.53 & 21.95 & 20.84 & 26.08 \\
Lebanese & \best{10.86} & 11.76 & 15.52 & 18.20 & 17.46 & 18.35 & 13.82 & 16.88 & 22.01 & 25.79 & 24.81 & 39.10 \\
\bottomrule
\end{tabular}}
\caption{CER (\%) by variety for all twelve systems; systems as in Table~\ref{tab:asr-main}.}
\label{tab:app-variety-cer}
\end{table}

\subsection{Public Test Sets: CER}
\label{app:asr-public-current}

Table~\ref{tab:app-public-current} gives \asr{}'s WER and CER on each of the
six public test sets under the leaderboard's current scoring
(Section~\ref{sec:asr-public}).

\begin{table}[H]
\centering
\footnotesize
\setlength{\tabcolsep}{4pt}
\begin{tabular}{lrrrrrrr}
\toprule
 & SADA & CV 18.0 & MASC-c & MASC-n & Casab. & MGB-2 & Avg \\
\midrule
WER & 25.76 & 3.95 & 5.35 & 16.70 & 41.72 & 9.56 & 17.17 \\
CER & 12.07 & 1.19 & 2.13 & 6.65 & 15.41 & 4.82 & 7.05 \\
\bottomrule
\end{tabular}
\caption{\asr{} on the six public test sets under the leaderboard's current
scoring (\%); Avg is the unweighted mean.}
\label{tab:app-public-current}
\end{table}